\documentclass{article}

\PassOptionsToPackage{numbers, compress}{natbib}

    \usepackage[preprint]{neurips_2025}

\usepackage[utf8]{inputenc} 
\usepackage[T1]{fontenc}    
\usepackage{hyperref}       
\usepackage{url}            
\usepackage{booktabs}       
\usepackage{amsfonts}       
\usepackage{nicefrac}       
\usepackage{microtype}      
\usepackage{xcolor}         
\usepackage{amsmath}
\usepackage{enumitem}
\usepackage{graphicx}

\usepackage{multirow}
\usepackage[utf8]{inputenc} 
\usepackage[T1]{fontenc}    
\usepackage{hyperref}       
\usepackage{url}            
\usepackage{booktabs}       
\usepackage{amsfonts}       
\usepackage{nicefrac}       
\usepackage{microtype}      
\usepackage{xcolor}         
\usepackage{amsmath}
\usepackage{enumitem}
\usepackage{xcolor}

\usepackage{multirow}
\usepackage{algorithm}
\usepackage{algpseudocode}
\usepackage{wrapfig}
\usepackage{times}
\usepackage{latexsym}
\usepackage{booktabs}
\usepackage{multirow}
\usepackage{makecell}
\usepackage{bm}
\usepackage{tabularray}
\usepackage{listings}
\usepackage{upquote}
\usepackage{amsfonts}
\usepackage{colortbl} 
\usepackage{graphicx}
\usepackage{amsfonts}  

\title{ Stable Reinforcement Learning for Efficient Reasoning}

\author{
    Muzhi Dai$^{\diamondsuit}$\footnotemark[1]~,
    Shixuan Liu$^{\heartsuit}$\thanks{$\quad$ Equal Contribution.}~,
    Qingyi Si$^{\diamondsuit}$\thanks{$\quad$ Corresponding Author.} ,
    \\
    $^\diamondsuit$Huawei Technologies Co., Ltd. 
    $^\heartsuit$Australian National University \\
    \fontsize{10.2pt}{0.1\baselineskip}\selectfont \texttt{mzdai666@gmail.com, u6920173@anu.edu.au, {siqingyi}@huawei.com}
}

\begin{document}

\maketitle

\begin{figure*}[h]
  \centering
  \includegraphics[width=\linewidth]{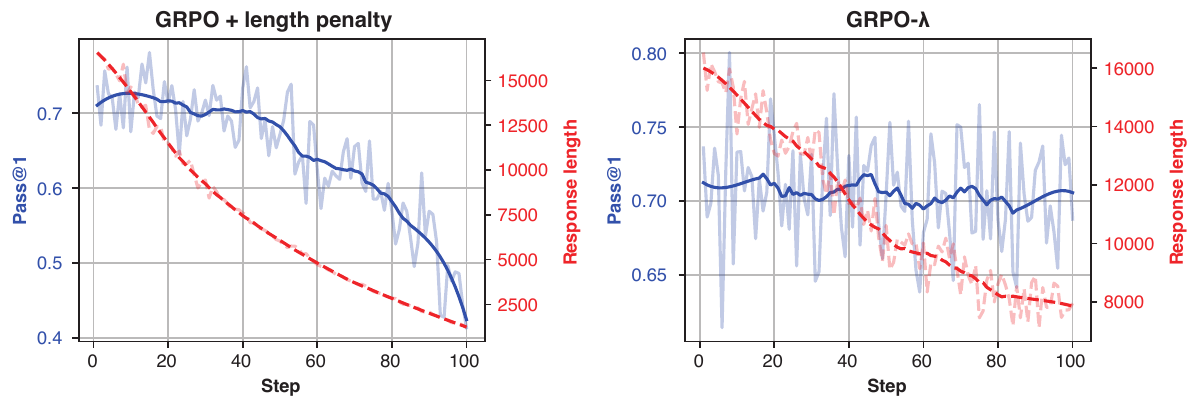}
  \caption{Training process of \textit{GRPO+length penalty} and our GRPO-$\lambda$.} 
  \label{fig1}
\end{figure*}

\begin{abstract}
 The success of Deepseek-R1 has drawn the LLM community's attention to reinforcement learning (RL) methods like GRPO. However, such rule-based  0/1 outcome reward methods lack the capability to regulate the intermediate reasoning processes during chain-of-thought (CoT)  generation, leading to severe overthinking phenomena. In response,   recent studies have designed reward functions to reinforce models' behaviors in producing shorter yet correct completions. Nevertheless, we observe that these length-penalty reward functions exacerbate RL training instability: as the completion length decreases, model accuracy abruptly collapses, often occurring early in training.  To address this issue, we propose a simple yet effective solution \textbf{GRPO-$\lambda$}, an efficient and stabilized variant of GRPO, which dynamically adjusts the reward strategy by monitoring the correctness ratio among completions within each query-sampled group. A low correctness ratio indicates the need to avoid length penalty that compromises CoT quality, triggering a switch to length-agnostic 0/1 rewards that prioritize reasoning capability. A high ratio maintains length penalties to boost efficiency. Experimental results show that our approach avoids training instability caused by length penalty while maintaining the optimal accuracy-efficiency trade-off. On the GSM8K, GPQA, MATH-500, AMC 2023, and AIME 2024 benchmarks, it improves average accuracy by 1.48\% while reducing CoT sequence length by 47.3\%.

\end{abstract}

\section{Introduction}

Recent advances in large language model (LLM) community have been driven by the development of test-time scaling \citep{snell2024scalingllmtesttimecompute}, demonstrating a positive correlation between generation length and models' reasoning capability, which is more effective than model-parameter scaling law \citep{kaplan2020scalinglawsneurallanguage}.  The open-source releases of DeepSeek-R1 \citep{deepseekai2025deepseekr1incentivizingreasoningcapability} and Qwen3 \citep{yang2025qwen3} have further stimulated recent research on reinforcement learning (RL) \citep{shao2024deepseekmathpushinglimitsmathematical,bai2022training,ouyang2022training,schulman2017proximal, ramesh2024group} for achieving reasoning models \citep{xu2025largereasoningmodelssurvey}. These models typically generate extended chain-of-thought (CoT) \citep{wei2023chainofthoughtpromptingelicitsreasoning} sequences containing rich and diverse reasoning paths.

However, recent studies \citep{chen2025think23overthinkingo1like,team2025kimi} have revealed that reasoning models often suffer from severe overthinking \citep{chen2025think23overthinkingo1like, cuadron2025danger} issues, characterized by excessive shallow reasoning steps and frequent thought-switching in prolonged CoTs \citep{wu2025more, cuadron2025danger,yang2025dynamicearlyexitreasoning}. This occurs because the rule-based outcome rewards in GRPO \citep{shao2024deepseekmathpushinglimitsmathematical} cannot effectively regulate intermediate reasoning processes. While longer reasoning chains statistically increase the probability of containing correct reasoning steps (thus improving answer accuracy and rewards during RL training), this GRPO mechanism continuously reinforces the lengthy CoT generation, and results in overthinking problems.


To address this issue, representative reasoning models like Kimi-1.5 \cite{kimiteam2025kimik15scalingreinforcement,arora2025traininglanguagemodelsreason,dai2025sgrpoearlyexitreinforcement} incorporate length penalty into RL training, constraining the model to generate higher-quality reasoning within shorter sequences, thereby mitigating overthinking while improving inference efficiency. For example, \cite{arora2025traininglanguagemodelsreason} assigns the highest reward to the shortest correct completion within the group. However, as shown in Figure 1 (left), we reveal that introducing length-aware reward or penalty functions leads to premature RL training collapse:  although CoT sequence length decreases as intended, model accuracy abruptly plummets, preventing stable RL training for sufficient iterations.


Intuitively, reasoning models require distinct training priorities at different competency stages: when reasoning capability is underdeveloped, reinforcement should prioritize accuracy, whereas efficiency optimization (via length penalty) should only be introduced once the model demonstrates sufficient reasoning capability. 
Current methods \citep{dai2025sgrpoearlyexitreinforcement,arora2025traininglanguagemodelsreason} overlook this progression, indiscriminately shortening CoT sequences for all samples during RL training, ultimately degrading the model's inherent reasoning capacity and causing RL training to collapse. Motivated by these insights, we propose a simple yet effective modification to GRPO, namely \textbf{GRPO-$\lambda$},  that sustainably improves reasoning efficiency without compromising reasoning accuracy, thereby preventing RL training collapse and ensuring sufficient training iterations, as shown in Figure 1(right). Specifically, we sample a set of completions per query following standard GRPO method, then evaluate the group-wise correctness rate, and dynamically switches between optimization modes: applying length penalties once correctness is adequately high (indicating mature reasoning capability to prioritize efficiency) or defaulting to standard GRPO's 0/1 outcome rewards (to reinforce accuracy fundamentals when below threshold). In this way, our method enables the joint optimization of reasoning efficiency and accuracy while ensuring training stability.


Experimental results on GSM8k \citep{cobbe2021trainingverifierssolvemathgsm8k}, GPQA \citep{rein2023gpqagraduatelevelgoogleproofqa}, AIME 2024 \citep{aime}, AMC 2023 \citep{AMC2023}, and MATH-500 \citep{math500hendrycks2021measuringmathematicalproblemsolving} benchmarks demonstrate that GRPO-$\lambda$ approach achieves the dual benefit: (1) enhanced training stability (enabling at least 2.5× more viable iterations) and (2) optimal performance-length tradeoffs, with a remarkable 47.3\% reduction in sequence length while improving accuracy by 1.48\%.

 \section{Related Work}
 The success of OpenAI-o1 \citep{openai2025learning,ouyang2022traininglanguagemodelsfollow} reveals that post-training through reinforcement learning serves as a mainstream paradigm for unlocking advanced reasoning capabilities in LLMs. Following the pioneering work of Deepseek-R1 \citep{deepseekai2025deepseekr1incentivizingreasoningcapability} and Qwen3 \citep{yang2025qwen3}, rule-based outcome reward RL methods \citep{deepseekai2025deepseekr1incentivizingreasoningcapability,kimiteam2025kimik15scalingreinforcement,gao2024designingeffectiverlreward,lambert2025tulu3pushingfrontiers,zeng2025simplerlzooinvestigatingtamingzero,wen2025lightr1curriculumsftdpo,song2025fastcurlcurriculumreinforcementlearning} like GRPO \citep{shao2024deepseekmathpushinglimitsmathematical} are widely adopted in post-training, encouraging models to produce long CoT outputs, at the cost of inducing overthinking issues \citep{chen2025think23overthinkingo1like, team2025kimi, cuadron2025danger}.  
 
 To solve it, recent studies \citep{dai2025sgrpoearlyexitreinforcement,arora2025traininglanguagemodelsreason,kimiteam2025kimik15scalingreinforcement} have independently proposed various length penalty mechanisms in reward function design. While these approaches share the common objective of promoting shorter responses and penalizing longer responses among correct ones, they implement distinct strategies. Specifically, Kimi 1.5 \citep{chen2025think23overthinkingo1like} first normalizes the length of sampled responses. For all responses exceeding 0.5 of the normalized length threshold, it assigns negative rewards, whereas those below receive positive rewards. Incorrect responses are restricted to a maximum reward of 0. Similarly, \cite{arora2025traininglanguagemodelsreason} employs a soft-clip sigmoid function to standardize and smooth length deviations from the group distribution.  This maps rewards to the interval (0,1), where shorter and correct responses receive values closer to 1, while incorrect responses are assigned zero reward. S-GRPO \citep{dai2025sgrpoearlyexitreinforcement} adopts a dual-rollout strategy, performing early-exit interventions at different positions within the first rollout response to construct a serial group, and allocating exponentially decaying rewards based on positional precedence, with zero reward for incorrect ones. 

Empirical observations reveal that such methods consistently induce premature collapse during RL training. This stems from their unilateral emphasis on length penalization without assessing potential compromises to the model's reasoning capability. In essence, when the model demonstrates strong pass@1 performance, length optimization should take priority for enhanced efficiency. Conversely, when sampled responses within the group fail to achieve satisfactory accuracy (weak pass@1), the focus should shift to reinforcing reasoning abilities rather than pursuing reasoning efficiently. Our method GRPO-$\lambda$ addresses this limitation by adaptively balancing these objectives, enabling more stable and prolonged RL training that ultimately achieves superior performance-efficiency trade-offs.

\section{Methods}

\begin{figure*}[t]
  \centering
  \includegraphics[width=\linewidth]{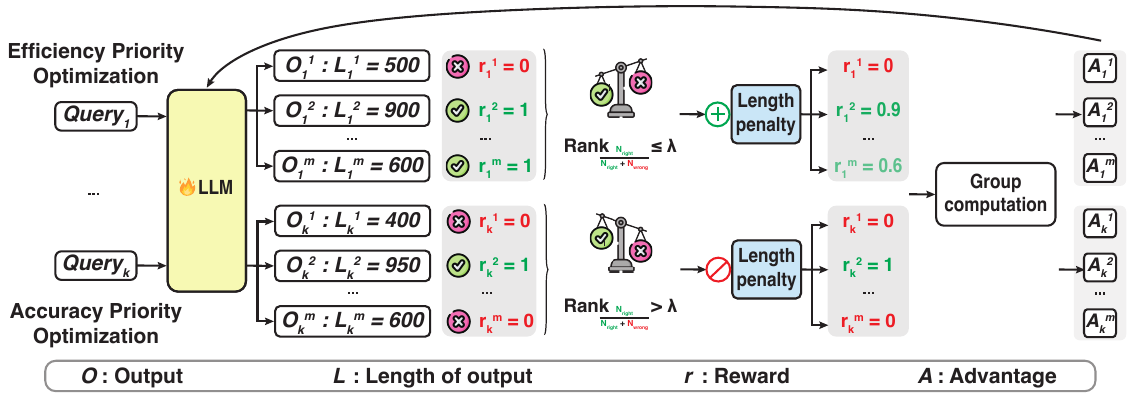}
  \caption{Framework of GRPO-$\lambda$.} 
  \label{fig2}
\end{figure*}

We introduce GRPO-$\lambda$, a stabilized and efficient variant of GRPO designed to address training instability caused by length-penalty reward. GRPO-$\lambda$ uses batch-wise dynamic adjustment of reward strategies, which selectively applies efficiency-prioritized or accuracy-prioritized optimization for different subsets of groups within a batch. This design ensures a controlled reduction in reasoning sequence length while maintaining accuracy, thereby preventing abrupt training collapse. Below, we detail the components and workflow of GRPO-$\lambda$.

\paragraph{Query-Sampled Group Generation.}
For each training query $Q_k$ in the batch, the model generates $m$ candidate completions $\{O_k^1, O_k^2, \dots, O_k^m\}$ using standard sampling techniques. Each completion $O_k^i$ is associated with: (1) Length $L_k^i$, indicating the number of tokens in the completion, and (2) Outcome Reward $r_k^i$, a binary 0/1 reward indicating whether $O_k^i$ is correct ($r_k^i = 1$) or incorrect ($r_k^i = 0$).

\paragraph{Batch-Wise Top-$\lambda$ Selection.}
For each batch of queries, we evaluate the correctness of each query-completion group and compute its correctness ratio. GRPO-$\lambda$ selects the top-$\lambda$ fraction of query-completion groups in terms of correctness ratio within the batch for efficiency-prioritized optimization. Specifically, the groups are ranked based on their correctness ratio within the batch. The top-$\lambda$ fraction (e.g., the top 20\%) is selected for efficiency-prioritized optimization, as shown in Figure 2 (Upper), as these groups demonstrate sufficient reasoning capability to focus on length reduction.  The remaining groups in the batch are assigned to accuracy-prioritized optimization to ensure that the model continues to improve its reasoning capability.



\paragraph{Dynamic Reward Strategy Adjustment.}
Based on the batch-wise top-$\lambda$ selection, GRPO-$\lambda$ applies two distinct reward strategies:

\begin{itemize}[left=0pt]
    \item Efficiency Priority Optimization (with Length Penalty):
For the top-$\lambda$ fraction of query-completion groups (those with higher correctness ratio), a length-penalty reward is applied to encourage shorter reasoning sequences:
\begin{equation}
r_{k}^{i} = 
\begin{cases}
1 - \alpha \cdot \sigma(\frac{L_{k}^{i} - \text{mean}(L_{k})_{\text{correct}}}{\text{std}(L_{k})_{\text{correct}}}) & \text{if $O_k^i$ is correct} \\
0 & \text{if $O_k^i$ is wrong}
\end{cases}
\end{equation}
where \(\alpha\) is the length penalty coefficient. \(\text{mean}(L_{k})_{\text{correct}}\) and \(\text{std}(L_{k})_{\text{correct}}\) are mean and standard deviation of completion lengths whose answers are correct, respectively. Incorrect completions ($r_k^i = 0$) receive no reward. This strategy prioritizes reasoning efficiency for groups that already demonstrate sufficient accuracy.

    \item Accuracy Priority Optimization (0/1 Outcome Reward):
For the remaining groups in the batch (those not in the top-$\lambda$ subset), the reward defaults to the standard GRPO 0/1 outcome reward:
\begin{equation}
r_{k}^{i} = 
\begin{cases}
1  & \text{if $O_k^i$ is correct} \\
0 & \text{if $O_k^i$ is wrong}
\end{cases}
\end{equation}
This strategy ensures that the model focuses on improving reasoning accuracy for completions with lower correctness scores.
\end{itemize}

This  reward strategy prevents the imbalanced emphasis on efficiency over accuracy that can arise from 
directly using length penalty for all groups \citep{kimiteam2025kimik15scalingreinforcement,arora2025training}. This ensures a controlled transition between accuracy and efficiency priorities, effectively curbing the risk of a sharp decline in accuracy. 


\paragraph{Advantage Computation and Parameter Update}
After obtaining the decaying rewards, like GRPO, GRPO-$\lambda$  calculates the advantage for each sample based on the group rewards. Specifically, the mean and standard deviation (\(\text{std}\)) of the rewards within the group are computed, and the advantage for each sample is calculated using the formula: $\hat{A}_{i} = \frac{r_i - \text{mean}(r_i)}{\text{std}(r_i)}$. Subsequently, the computed advantage for each sample is broadcast to all corresponding response tokens. Finally, parameter updates are performed based on the advantage values of each sample.


\newcommand{\annotate}[3]{%
    #1\raisebox{-0.5ex}{\scalebox{0.7}{\textcolor{#2}{#3}}}%
}

\begin{table}[t]
\centering
\scriptsize
\caption{ Experimental results on Qwen3-8B. "LP" indicates length penalty. * indicates results trained with identical step counts to GRPO-$\lambda$,  having undergone training collapse. "Acc" denotes accuracy,
"Tok" denotes token count, and "CR" denotes compression rate. The top-2 best results are in bold.}
\label{table2}
\setlength\tabcolsep{2pt} 
\renewcommand{\arraystretch}{1}
\begin{tabular}{@{}lccccccccccccccccc@{}} 
\toprule
 \multirow{2}{*}{\textbf{Method}} 
 & \multicolumn{3}{c}{\textbf{GSM8K}} & \multicolumn{3}{c}{\textbf{GPQA}} & \multicolumn{3}{|c}{\textbf{MATH-500}} & \multicolumn{3}{c}{\textbf{AMC 2023}} & \multicolumn{3}{c}{\textbf{AIME 2024}}   & \multicolumn{2}{|c}{\textbf{Overall}} \\
    & {Acc$\uparrow$} & {Tok$\downarrow$} & {CR$\downarrow$} & Acc$\uparrow$ & Tok$\downarrow$ & CR$\downarrow$ & {Acc$\uparrow$ } & {Tok$\downarrow$} & CR$\downarrow$ & {Acc$\downarrow$} & {Tok$\downarrow$} & CR$\downarrow$ & {Acc$\uparrow$ } & {Tok$\downarrow$} & CR$\downarrow$ & Acc$\uparrow$  & CR$\downarrow$ \\ 
\hline
\multicolumn{18}{l}{{\cellcolor[rgb]{0.957,0.957,0.957}}\textit{\textbf{Qwen3-8B}}} \\
\textit{Vanilla} & 95.4 & 2,370 & 100\% & 55.6 & 8,741 & 100\% & \multicolumn{1}{|c}{93.4} & 5,577 & 100\% & 91.3 & 9,452 & 100\% & 74.1 & 15,326 & 100\% & \multicolumn{1}{|l}{{81.90}} & 100\%   \\
\textit{+GRPO} & 95.8 & 2,355 & 99.4\% & 55.8 & 8,819 & 100.9\% & \multicolumn{1}{|c}{94.4} & 5,440 & 97.5\% & 92.8 & 8,983 & 95.0\% & 72.7 & 15,154 & 98.9\% & \multicolumn{1}{|l}{{\textbf{82.30}}} & 98.34\%  \\
\textit{\ \  +LP} & 95.4 & 1,323 & 55.8\% & 55.4 & 4,930 & 56.4\% & \multicolumn{1}{|c}{94.2} & 2,874 & 51.5\% & 92.8 & 4,933 & 52.2\% & 71.9 & 9,266 & 60.5\% & \multicolumn{1}{|l}{{81.94}} & 55.28\%  \\
\textit{\ \  +LP*} & 94.6 & 250 & 10.5\% & 53.8 & 732 & 8.4\% & \multicolumn{1}{|c}{86.0} & 507 & 9.1\% & 75.9 & 874 & 9.2\% & 32.1 & 2,037 & 13.3\% & \multicolumn{1}{|l}{{68.48}} & \textbf{10.1\%}  \\
\rowcolor[rgb]{0.9,0.95,1}
\textit{+\textbf{GRPO-$\lambda$}} & 95.5 & 1,114 & 47.0\% & 56.8 & 4,872 & 55.7\% & \multicolumn{1}{|c}{96.0} & 2,990 & 53.6\% & 94.4 & 4,751 & 50.3\% & 74.4 & 8,714 & 56.9\% & \multicolumn{1}{|l}{{\textbf{83.42}}} & \textbf{52.7}\%  \\
 \bottomrule
\end{tabular}
\end{table}

\section{Experiments}
\subsection{Benchmarks and Settings.}

We conducted comprehensive evaluations of our method on several mainstream reasoning benchmarks, including mathematical tasks (GSM8K \citep{cobbe2021trainingverifierssolvemathgsm8k}, MATH-500 \citep{math500hendrycks2021measuringmathematicalproblemsolving}, and the more challenging AMC 2023 \citep{AMC2023} and AIME 2024 \citep{aime}) as well as the scientific reasoning benchmark GPQA \citep{rein2023gpqagraduatelevelgoogleproofqa}.

We choose Qwen3-8B \cite{yang2025qwen3} as the base model for experiments. For training data, we select queries from DeepMath-103K \cite{he2025deepmath103klargescalechallengingdecontaminated}. Specifically, we sample 8 times for each query using Qwen3-8B, and select queries that can be answered correctly 2-6 times.  
During training, we use a learning rate of \(1 \times 10^{-6}\) and randomly sample 16 times for each query. The generation batch size and training batch size are both set to \(128 \times 16\). For the length penalty, we set the scalar parameter $\alpha$ to 0.2. For GRPO-$\lambda$, we set $\lambda$ equal to 20\%.  Across all experiments, we employ Adam \cite{kingma2014adam} as the standard optimizer.

\subsection{Experimental Results}

%
As shown in Table 1, our method achieves the optimal trade-off between accuracy and efficiency. Compared to the conventional \textit{GRPO+length penalty} approach, GRPO-$\lambda$ further improves the average accuracy by 1.48\% while achieving more significant sequence length compression on five benchmarks. Notably,  For more challenging mathematical tasks (e.g., AIME 2024, AMC 2023), the benefits of our method become even more pronounced, as the relatively simpler mathematical and scientific tasks (e.g., GSM8K, GPQA datasets)  are less sensitive to length variations. 
 
 The results of \textit{GRPO+length penalty*} confirm that incorporating length penalty into the reward function leads to training collapse. Specifically, when trained for the same number of steps as GRPO-$\lambda$ , \textit{GRPO+length penalty*} achieves more significant sequence length compression of 89.9\% but suffers a substantial accuracy drop of 13.42\%. This phenomenon should be avoided in post-training optimization, as length compression without preserving accuracy becomes meaningless. Furthermore, as shown in Figure \ref{fig1}, the accuracy of \textit{GRPO+length penalty} begins to decline after 40 steps, whereas our method maintains stable performance even at 100 steps, extending effective training steps by at least 2.5×. This demonstrates that our approach provides stable reinforcement learning for efficient reasoning.

\begin{wrapfigure}[20]{r}{0.6\textwidth}
  \centering
  \includegraphics[width=7.0cm]{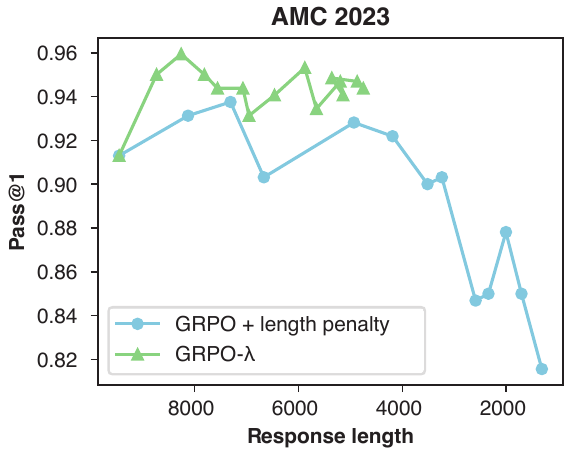}
  \caption{Relationship between performance and response length of GRPO + length penalty and GRPO-$\lambda$ on AMC 2023 benchmark as training progresses.}
  \label{fig3}
\end{wrapfigure}

\subsection{Discussion.}
%
 

Figure \ref{fig3} presents the relationship between CoT length and accuracy for \textit{GRPO+length penalty} and GRPO-$\lambda$, where our method's curve consistently occupies the Pareto-superior region to the left and above \textit{GRPO+length penalty}'s curve. Specifically, when \textit{GRPO+length penalty} attains similar lengths to our approach, we observe a significant accuracy gap in our favor; conversely, when matching our accuracy levels, \textit{GRPO+length penalty} requires substantially longer reasoning chains (e.g., $\sim$ 7000 vs. $\sim$ 5000 tokens at accuracy $\approx$ 0.94). 
 
 As the sequence length progressively decreases, the accuracy of \textit{GRPO+length penalty} exhibits a consistent decline, whereas our method maintains robust stability in performance. Crucially, recent studies \citep{snell2024scalingllmtesttimecompute,hou2025thinkprunepruninglongchainofthought} reveal that excessive length reduction inevitably compromises the model's reasoning capability. GRPO-$\lambda$ adaptively optimizes sequence length within an appropriate range without sacrificing accuracy. Notably, the dense clustering of data points around the length of 5000 suggests this represents the minimal length preserving model accuracy, which serves as a critical threshold that our method automatically converges to. 

\begin{figure*}[t]
  \centering
  \includegraphics[width=\linewidth]{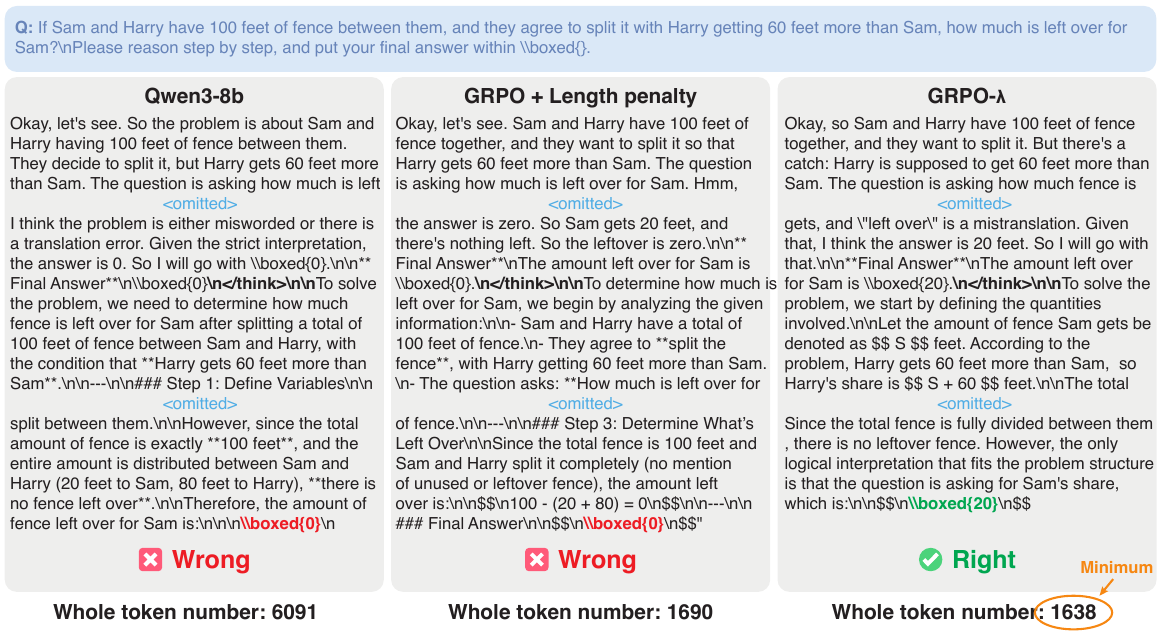}
  \caption{Comparison of a generated content sample on GSM8K.} 
  \label{fig4}
\end{figure*}

 Figure \ref{fig4} presents case samples that reveal three distinct behaviors: Qwen3-8b, while generating the longest response, provides incorrect answers due to its overthinking issue; \textit{GRPO+length penalty} successfully reduces sequence length but at the cost of impairing the model's reasoning capability, resulting in erroneous responses; in contrast, our method achieves correct answers while operating at the shortest sequence length. 
 

 \section{Conclusion and Future Work}


 This paper presents the first systematic study on how length-penalty reward design impacts RL training stability in post-training and proposes GRPO-$\lambda$, a simple yet effective method. Through extensive experiments, we reveal critical insights for balancing efficiency and accuracy. Specifically, the CoT length reduction rate must be carefully controlled, as excessively rapid shortening inevitably degrades accuracy. Evaluations on the GSM8K, GPQA, MATH-500, AMC 2023, and AIME 2024 benchmarks demonstrate that our method achieves a superior accuracy-efficiency trade-off (+1.48\% accuracy with 47.3\% shorter CoT) and enhances training stability for RL of efficient reasoning.


 During our experimental exploration, we made several critical observations: (1) Overly aggressive length reduction during training causes premature reduction of reasoning paths before the model properly adjusts them, thereby impairing the exploration of reasoning processes and ultimately hurting accuracy. (2) The difficulty level of training data proves crucial, as oversimplified data lead to rapid collapse of chain-of-thought length. (3) The proportion of length-penalty groups in each batch ($\lambda$ value) significantly impacts performance, where too large proportion makes accuracy difficult to maintain. These insights will guide our comprehensive empirical study in the future version through systematic experiments addressing all three aspects.

Beyond these findings, our methodology's core principles suggest promising extensions. For instance, when the model approaches a critical length reduction threshold near performance collapse, timely intervention could be implemented by training with GRPO at a proper setting of max length for stabilization, potentially enabling accuracy improvements while maintaining the compressed length.


\bibliographystyle{unsrtnat}
\bibliography{neurips_2025}


\end{document}